\documentclass[11pt,letterpaper]{article}
\pdfoutput=1
\usepackage{emnlp2017}
\usepackage{times}
\usepackage{url}
\usepackage{latexsym}

\usepackage{mathtools}
\usepackage{calc}
\usepackage{amsmath, amsthm, amssymb, amsfonts}
\emnlpfinalcopy

\DeclareMathOperator*{\argmax}{arg\,max}

\title{Dependency Parsing with Dilated Iterated Graph CNNs}

\author{Emma Strubell \qquad Andrew McCallum\\
  College of Information and Computer Sciences \\
  University of Massachusetts Amherst \\
  {\tt \{strubell, mccallum\}@cs.umass.edu}
}

\date{}

\begin{document}
\maketitle
\begin{abstract}

Dependency parses are an effective way to inject linguistic knowledge into many downstream tasks, and
many practitioners wish to efficiently parse sentences at scale. 
Recent advances in GPU hardware have enabled neural networks to achieve significant gains over the previous best models, these models still fail to leverage GPUs' capability for massive parallelism due to their requirement of sequential processing of the sentence. In response, we propose Dilated Iterated Graph Convolutional Neural Networks (DIG-CNNs) for
graph-based dependency parsing, a graph convolutional architecture that allows for efficient end-to-end GPU parsing. In experiments on the English Penn TreeBank benchmark, we show that DIG-CNNs perform on par with some of the best neural network parsers. 
\end{abstract}

\section{Introduction}

By vastly accelerating and parallelizing the core numeric operations for performing inference and computing gradients in neural networks, recent developments in GPU hardware have facilitated the emergence of deep neural networks as state-of-the-art models for many NLP tasks, such as syntactic dependency parsing. The best neural dependency parsers generally consist of two stages: First, they employ a recurrent neural network such as a bidirectional LSTM to encode each token in context; next, they compose these token representations into a parse tree. Transition based dependency parsers \citep{nivre2009non, chen2014fast, andor2016globally} produce a well-formed tree by predicting and executing a series of shift-reduce actions, whereas graph-based parsers \citep{mcdonal2005non,kiperwasser2016simple, dozat2017deep} generally employ attention to produce marginals over each possible edge in the graph, followed by a dynamic programming algorithm to find the most likely tree given those marginals. 

\begin{figure}
\hspace{-0.5cm}
\includegraphics[scale=0.48]{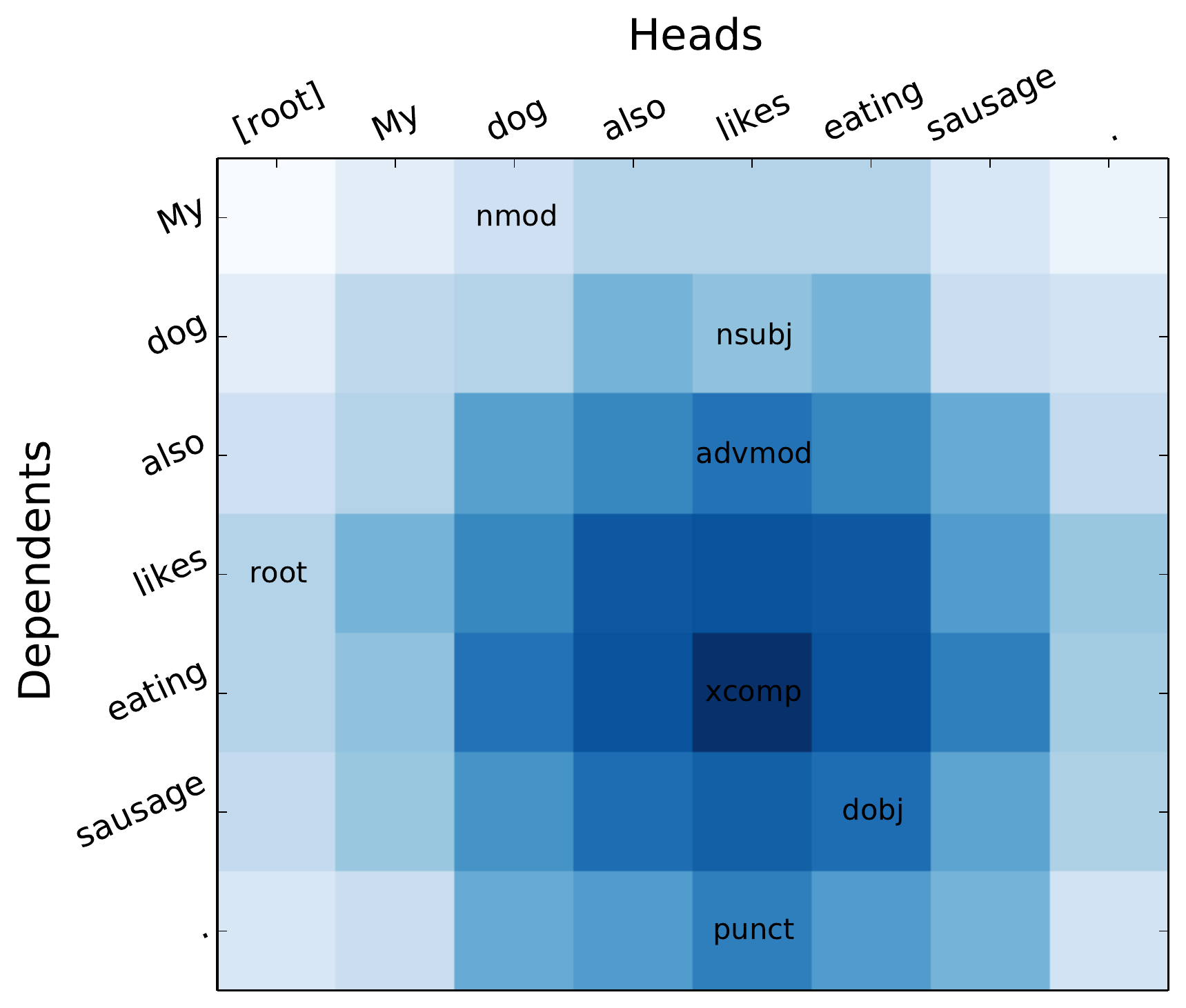}
\caption{Receptive field for predicting the head-dependent relationship between \emph{likes} and \emph{eating}. Darker cell indicates more layers include that cell's representation. Heads and labels corresponding to gold tree are indicated. \label{dilated-block-fig}}
\end{figure}

Because of their dependency on sequential processing of the sentence, none of these architectures fully exploit the massive parallel processing capability that GPUs possess. If we wish to maximize GPU resources, graph-based dependency parsers are more desirable than their transition-based counterparts since attention over the edge-factored graph can be parallelized across the entire sentence, unlike the transition-based parser which must sequentially predict and perform each transition. By encoding token-level representations with an Iterated Dilated CNN (ID-CNN) \citep{strubell2017fast}, we can also remove the sequential dependencies of the RNN layers. Unlike \citet{strubell2017fast} who use 1-dimensional convolutions over the sentence to produce token representations, our network employs 2-dimensional convolutions over the adjacency matrix of the sentence's parse tree, modeling attention from the bottom up. By training with an objective that encourages our model to predict trees using only simple matrix operations, we additionally remove the additional computational cost of dynamic programming inference. Combining all of these ideas, we present Dilated Iterated Graph CNNs (DIG-CNNs): a combined convolutional neural network architecture and training objective for efficient, end-to-end GPU graph-based dependency parsing.

We demonstrate the efficacy of these models in experiments on English Penn TreeBank, in which our models perform similarly to the state-of-the-art.

\section{Dilated Convolutions}

Though common in other areas such as computer vision, 2-dimensional convolutions are rarely used in NLP since it is usually unclear how to process text as a 2-dimensional grid. However, 2-dimensional convolutional layers are a natural model for embedding the adjacency matrix of a sentence's parse.

A 2-dimensional convolutional neural network layer transforms each input element, in our case an edge in the dependency graph, as a linear function of the width $r_w$ and height $r_h$ window of surrounding input elements (other possible edges in the dependency graph). In this work we assume square convolutional windows: $r_h = r_w$.

Dilated convolutions perform the same operation, except rather than transforming directly adjacent inputs, the convolution is defined over a wider input window by skipping over $\delta$ inputs at a time, where $\delta$ is the dilation width. A dilated convolution of width 1 is equivalent to a simple convolution. Using the same number of parameters as a simple convolution with the same radius, the $\delta > 1$ dilated convolution incorporates broader context into the representation of a token than a simple convolution.

\subsection{Iterated Dilated CNNs}
\label{blocks-section}
Stacking many dilated CNN layers can easily incorporate information from a whole sentence. For example, with a radius of 1 and 4 layers of dilated convolutions, the effective input window size for each token is width 31, which exceeds the average sentence length (23) in the Penn TreeBank corpus. However, simply increasing the depth of the CNN can cause considerable over-fitting when data is sparse relative to the growth in model parameters. To address this, we employ Iterated Dilated CNNs (ID-CNNs) \citep{strubell2017fast}, which instead apply the same small stack of dilated convolutions repeatedly, each time taking the result of the last stack as input to the current iteration. Applying the parameters recurrently in this way increases the size of the window of context incorporated into each token representation while allowing the model to generalize well. Their training objective additionally computes a loss for the output of each application, encouraging parameters that allow subsequent stacks to resolve dependency violations from their predecessors.

\section{Dilated Iterated Graph CNNs}
\label{blocks-section}

We describe how to extend ID-CNNs \citep{strubell2017fast} to 2-dimensional convolutions over the adjacency matrix of a sentence's parse tree, allowing us to model the parse tree through the whole network, incorporating evidence about nearby head-dependent relationships in every layer of the network, rather than modeling at the token level followed by a single layer of attention to produce head-dependent compatibilities between tokens. ID-CNNs allow us to efficiently incorporate evidence from the entire tree without sacrificing generalizability.

\subsection{Model architecture\label{model-section}}
Let $x = [x_1, \ldots, x_T]$ be our input text\footnote{In practice, we include a dummy root token at the beginning of the sentence which serves as the head of the root. We do not predict a head for this dummy token.} Let $y = [y_{1}, \ldots, y_{T}]$ be labels with domain size $D$ for the edge between each token $x_i$ and its head $x_j$. We predict the most likely $y$, given a conditional model $P(y | x)$ where the tags are conditionally independent given some features for $x$:
\begin{equation}
P(y | x) = \prod_{t = 1}^{T} P(y_t | F(x)),
\label{eq:cond-ind}
\end{equation}

The local conditional distributions of Eqn.~\eqref{eq:cond-ind} come from a straightforward extension of ID-CNNs \citep{strubell2017fast} to 2-dimensional convolutions. This network takes as input a sequence of $T$ vectors ${\bf x_t}$, and outputs a $T\times T$ matrix of per-class scores ${\bf h_{ij}}$ for each pair of tokens in the sentence. 

We denote the $k$th dilated convolutional layer of dilation width $\delta$ as $D_\delta^{(k)}$. The first layer in the network transforms the input to a graph by concatenating all pairs of vectors in the sequence ${\bf x_i}, {\bf x_j}$ and applying a 2-dimensional dilation-1 convolution $D_1^{(0)}$ to form an initial edge representation ${\bf c_{ij}^{(0)}}$ for each token pair:
\begin{align}
{\bf c_{ij}}^{(0)} = D_1^{(0)}{[{\bf x_i};{\bf x_j}]}
\end{align}
We denote vector concatenation with $[\cdot;\cdot]$. Next, $L_c$ layers of dilated convolutions of exponentially increasing dilation width are applied to ${\bf c_{ij}}^{(0)}$, folding in increasingly broader context into the embedded representation of ${\bf e_{ij}}$ at each layer. Let $r()$ denote the ReLU activation function \citep{glorot2011deep}. Beginning with ${\bf c_t}^{(0)} = {\bf i_t}$ we define the stack of layers with the following recurrence:
\begin{align}
{\bf c_{ij}}^{(k)} &= r\left(D_{2^{L_c-1}}^{(k-1)}{\bf c_t}^{(k-1)}\right)
\end{align}
and add a final dilation-1 layer to the stack:
\begin{align}
{\bf c_{ij}}^{(L_c+1)} &= r\left(D_1^{(L_c)}{\bf c_t}^{(L_c)}\right)
\end{align}
We refer to this stack of dilated convolutions as a \emph{block} $B(\cdot)$, which has output resolution equal to its input resolution. To incorporate even broader context without over-fitting, we avoid making $B$ deeper, and instead iteratively apply $B$ $L_b$ times, introducing no extra parameters. Starting with ${\bf b_t}^{(1)} = B\left({\bf i_t}\right)$, we define the output of block $m$:
\begin{align}
{\bf b_{ij}}^{(m)} &= B\left({\bf b_t}^{(m-1)} \right)
\label{block-eqn}
\end{align}
We apply a simple affine transformation $W_o$ to this final representation to obtain label scores for each edge ${\bf e_{ij}}$:
\begin{align}
{\bf h_{ij}}^{(L_b)} = W_o{\bf b_t}^{(L_b)}
\label{outputs-eqn}
\end{align}

We can obtain the most likely head (and its label) for each dependent by computing the argmax over all labels for all heads for each dependent:
\begin{align}
{\bf h_t} = \argmax_{j}{\bf h_{ij}}^{(L_b)}
\end{align}




\subsection{Training}

Our main focus is to apply the DIG-CNN as feature extraction for the conditional model described in Sec.~\ref{model-section}, where tags are conditionally independent given deep features, since this will enable prediction that is parallelizable across all possible edges. Here, maximum likelihood training is straightforward because the likelihood decouples into the sum of the likelihoods of independent logistic regression problems for every edge, with natural parameters given by Eqn.~\eqref{outputs-eqn}:
\begin{equation}
\frac{1}{T}\sum_{t=1}^T \log P(y_t \mid {\bf h_{t}}) \label{eq:loss}
\end{equation}

We could also use the DIG-CNN as input features for an MST parser, where the partition function and its gradient are computed using Kirchhoff’s Matrix-Tree Theorem \citep{tutte1984graph}, but our aim is to approximate inference in a tree-structured graphical model using greedy inference and expressive features over the input in order to perform inference as efficiently as possible on a GPU. 

To help bridge the gap between these two techniques, we use the training technique described in \citep{strubell2017fast}. The tree-structured graphical model has preferable sample complexity and accuracy since prediction directly reasons in the space of structured outputs. Instead, we compile some of this reasoning in output space into DIG-CNN feature extraction. Instead of explicit reasoning over output labels during inference, we train the network such that each block is predictive of output labels. Subsequent blocks learn to correct dependency violations of their predecessors, refining the final sequence prediction.

To do so, we first define predictions of the model after each of the $L_b$ applications of the block. Let ${\bf h_{t}}^{(m)}$ be the result of applying the matrix $W_o$ from~\eqref{outputs-eqn} to ${\bf b_{t}}^{(m)}$, the output of block $m$. We minimize the average of the losses for each application of the block: 
\begin{align}
\frac{1}{L_b}\sum_{k=1}^{L_b} \frac{1}{T}\sum_{t=1}^T \log P(y_t \mid {\bf h_{t}}^{(m)}). \label{eq:avg-loss}
\end{align}

By rewarding accurate predictions after each application of the block, we learn a model where later blocks are used to refine initial predictions. The loss also helps reduce the vanishing gradient problem~\citep{hochreiter1998vanishing} for deep architectures. 




We apply dropout~\citep{srivastava2014dropout} to the raw inputs ${\bf x_{ij}}$ and to each block's output ${\bf b_t}^{(m)}$ to help prevent overfitting.

\section{Related work\label{related-work}}

Currently, the most accurate parser in terms of labeled and unlabeled attachment scores is the neural network graph-based dependency parser of \citet{dozat2017deep}. Their parser builds token representations with a bidirectional LSTM over word embeddings, followed by head and dependent MLPs. Compatibility between heads and dependents is then scored using a biaffine model, and the highest scoring head for each dependent is selected. 

Previously, \citep{chen2014fast} pioneered neural network paring with a transition-based dependency parser which used features from a fast feed-forward neural network over word, token and label embeddings. Many improved upon this work by increasing the size of the network and using a structured training objective \citep{weiss2015structured,andor2016globally}. \citep{kiperwasser2016simple} were the first to present a graph-based neural network parser, employing an MLP with bidirectional LSTM inputs to score arcs and labels. \citep{cheng2016bidirectional} propose a similar network, except with additional forward and backward encoders to allow for conditioning on previous predictions. \citep{kuncoro2016distilling} take a different approach, distilling a consensus of many LSTM-based transition-based parsers into one graph-based parser. \citep{ma2017neural} employ a similar model, but add a CNN over characters as an additional word representation and perform structured training using the Matrix-Tree Theorem. \citet{hashimoto2017joint} train a large network which performs many NLP tasks including part-of-speech tagging, chunking, graph-based parsing, and entailment, observing benefits from multitasking with these tasks.

Despite their success in the area of computer vision, in NLP convolutional neural networks have mainly been relegated to tasks such as sentence classification, where each input sequence is mapped to a single label (rather than a label for each token)~\citet{kim2014convolutional,kalchbrenner2014convolutional,zhang2015character,toutanova2015representing}. As described above, CNNs have also been used to encode token representations from embeddings of their characters, which similarly perform a pooling operation over characters. \citet{lei2015molding} present a CNN variant where convolutions adaptively skip neighboring words. While the flexibility of this model is powerful, its adaptive behavior is not well-suited to GPU acceleration.

More recently, inspired by the success of deep dilated CNNs for image segmentation in computer vision \citep{yu2015multi,chen2015semantic}, convolutional neural networks have been employed as fast models for tagging, speech generation and machine translation. \citep{vandenoord2016wavenet} use dilated CNNs to efficiently generate speech, and \citet{kalchbrenner2016neural} describes an encoder-decoder model for machine translation which uses dilated CNNs over bytes in both the encoder and decoder. \citet{strubell2017fast} first described the one-dimensional ID-CNN architecture which is the basis for this work, demonstrating its success as a fast and accurate NER tagger. \citet{gehring2017convolutional} report state-of-the-art results and much faster training from using many CNN layers with gated activations as encoders and decoders for a sequence-to-sequence model. While our architecture is similar to the encoder architecture of these models, ours is differentiated by (1) being tailored to smaller-data regimes such as parsing via our iterated architecture and loss, and (2) employing two-dimensional convolutions to model the adjacency matrix of the parse tree. We are the first to our knowledge to use dilated convolutions for parsing, or to use two-dimensional dilated convolutions for NLP.

\section{Experimental Results}

\subsection{Data and Evaluation}
We train our parser on the English Penn TreeBank on the typical data split: training on sections 2--21, testing on section 23 and using section 22 for development. We convert constituency trees to dependencies using the Stanford dependency framework v3.5 \cite{deMarneffe2008}, and use part-of-speech tags from the Stanford left3words part-of-speech tagger. As is the norm for this dataset, our evaluation excludes punctuation. Hyperparameters that resulted in the best performance on the validation set were selected via grid search. A more detailed description of optimization and data pre-processing can be found in the Appendix.

\subsection{English PTB Results}

We compare our models labeled and unlabeled attachment scores to the neural network graph-based dependency parsers described in Sec. \ref{related-work}. Without enforcing trees at test time, our model performs just under the LSTM-based parser of \citet{kiperwasser2016simple}, and a few points lower than the state-of-the-art. When we post-process our model's outputs into trees, like all the other models in our table, our results increase to perform slightly above \citet{kiperwasser2016simple}.

\begin{table}
\begin{tabular}{lll}
    Model & UAS & LAS \\ \hline \hline
    \citet{kiperwasser2016simple} & 93.9 & 91.9 \\ 
    \citet{cheng2016bidirectional} & 94.10 & 91.49 \\ 
    \citet{kuncoro2016distilling} & 94.3 & 92.1 \\
    \citet{hashimoto2017joint} & 94.67 & 92.90 \\
    \citet{ma2017neural} & 94.9 & 93.0 \\
    \citet{dozat2017deep} & 95.74 & 94.08 \\ \hline 
    DIG-CNN & 93.70 & 91.72 \\
    DIG-CNN + Eisner & 94.03 & 92.00 \\ 
  \end{tabular}
  \caption{Labeled and unlabeled attachment scores of our model compared to state-of-the-art graph-based parsers \label{results-table}}
\end{table}

We believe our model's relatively poor performance compared to existing models is due to its limited incorporation of context from the entire sentence. While each bidirectional LSTM token representation observes all tokens in the sentence, our reported model observes a relatively small window, only 9 tokens. We hypothesize that this window is not sufficient for producing accurate parses. Still, we believe this is a promising architecture for graph-based parsing, and with further experimentation could meet or exceed the state-of-the-art while running faster by better leveraging GPU architecture.

\section{Conclusion}

We present DIG-CNNs, a fast, end-to-end convolutional architecture for graph-based dependency parsing. Future work will experiment with deeper CNN architectures which incorporate broader sentence context in order to increase accuracy without sacrificing speed.

\section*{Acknowledgments}
We thank Patrick Verga and David Belanger for helpful discussions. This work was supported in part by the Center for Intelligent Information Retrieval, in part by DARPA under agreement number FA8750-13-2-0020, in part by Defense Advanced Research Agency (DARPA) contract number HR0011-15-2-0036, in part by the National Science Foundation (NSF) grant number DMR-1534431, and in part by the National Science Foundation (NSF) grant number IIS-1514053. The U.S. Government is authorized to reproduce and distribute reprints for Governmental purposes notwithstanding any copyright notation thereon. Any opinions, findings and conclusions or recommendations expressed in this material are those of the authors and do not necessarily reflect those of the sponsor.

\bibliography{emnlp2017-ws}

\begin{thebibliography}{}
\expandafter\ifx\csname natexlab\endcsname\relax\def\natexlab#1{#1}\fi

\bibitem[{Andor et~al.(2016)Andor, Alberti, Weiss, Severyn, Presta, Ganchev,
  Petrov, and Collins}]{andor2016globally}
Daniel Andor, Chris Alberti, David Weiss, Aliaksei Severyn, Alessandro Presta,
  Kuzman Ganchev, Slav Petrov, and Michael Collins. 2016.
\newblock Globally normalized transition-based neural networks.
\newblock In {\em Proceedings of the 54th Annual Meeting of the Association for
  Computational Linguistics\/}.

\bibitem[{Chen and Manning(2014)}]{chen2014fast}
Danqi Chen and Christopher~D. Manning. 2014.
\newblock A fast and accurate dependency parser using neural networks.
\newblock In {\em EMNLP\/}.

\bibitem[{Chen et~al.(2015)Chen, Papandreou, Kokkinos, Murphy, and
  Yuille}]{chen2015semantic}
Liang-Chieh Chen, George Papandreou, Iasonas Kokkinos, Kevin Murphy, and
  Alan~L. Yuille. 2015.
\newblock Semantic image segmentation with deep convolutional nets and fully
  connected crfs.
\newblock In {\em ICLR\/}.

\bibitem[{Cheng et~al.(2016)Cheng, Fang, He, Gao, and
  Deng}]{cheng2016bidirectional}
Hao Cheng, Hao Fang, Xiaodong He, Jianfeng Gao, and Li~Deng. 2016.
\newblock Bi-directional attention with agreement for dependency parsing.
\newblock In {\em EMNLP\/}.

\bibitem[{de~Marneffe and Manning(2008)}]{deMarneffe2008}
Marie-Catherine de~Marneffe and Christopher~D. Manning. 2008.
\newblock The stanford typed dependencies representation.
\newblock In {\em COLING 2008 Workshop on Cross-framework and Cross-domain
  Parser Evaluation\/}.

\bibitem[{Dozat and Manning(2017)}]{dozat2017deep}
Timothy Dozat and Christopher~D. Manning. 2017.
\newblock Deep biaffine attention for neural dependency parsing.
\newblock In {\em ICLR\/}.

\bibitem[{Gehring et~al.(2017)Gehring, Auli, Grangier, Yarats, and
  Dauphin}]{gehring2017convolutional}
Jonas Gehring, Michael Auli, David Grangier, Denis Yarats, and Yann~N. Dauphin.
  2017.
\newblock Convolutional sequence to sequence learning.
\newblock {\em arXiv preprint: arXiv:1705.03122\/} .

\bibitem[{Glorot et~al.(2011)Glorot, Bordes, and Bengio}]{glorot2011deep}
Xavier Glorot, Antoine Bordes, and Yoshua Bengio. 2011.
\newblock Deep sparse rectifier neural networks.
\newblock In {\em AISTATS\/}.

\bibitem[{Hashimoto et~al.(2017)Hashimoto, Xiong, Tsuruoka, and
  Socher}]{hashimoto2017joint}
Kazuma Hashimoto, Caiming Xiong, Yoshimasa Tsuruoka, and Richard Socher. 2017.
\newblock A joint many-task model: Growing a neural network for multiple nlp
  tasks.
\newblock {\em arXiv preprint: arXiv:1611.01587\/} .

\bibitem[{Hochreiter(1998)}]{hochreiter1998vanishing}
Sepp Hochreiter. 1998.
\newblock The vanishing gradient problem during learning recurrent neural nets
  and problem solutions.
\newblock {\em International Journal of Uncertainty, Fuzziness and
  Knowledge-Based Systems\/} 6(02):107--116.

\bibitem[{Kalchbrenner et~al.(2016)Kalchbrenner, Espeholt, Simonyan, van~den
  Oord, Graves, and Kavukcuoglu}]{kalchbrenner2016neural}
Nal Kalchbrenner, Lasse Espeholt, Karen Simonyan, Aaron van~den Oord, Alex
  Graves, and Koray Kavukcuoglu. 2016.
\newblock Neural machine translation in linear time.
\newblock {\em arXiv preprint arXiv:1610.10099\/} .

\bibitem[{Kalchbrenner et~al.(2014)Kalchbrenner, Grefenstette, and
  Blunsom}]{kalchbrenner2014convolutional}
Nal Kalchbrenner, Edward Grefenstette, and Phil Blunsom. 2014.
\newblock A convolutional neural network for modelling sentences.
\newblock In {\em Proceedings of the 52nd Annual Meeting of the Association for
  Computational Linguistics\/}.

\bibitem[{Kim(2014)}]{kim2014convolutional}
Yoon Kim. 2014.
\newblock Convolutional neural networks for sentence classification.
\newblock In {\em EMNLP\/}.

\bibitem[{Kiperwasser and Goldberg(2016)}]{kiperwasser2016simple}
Eliyahu Kiperwasser and Yoav Goldberg. 2016.
\newblock Simple and accurate dependency parsing using bidirectional lstm
  feature representations.
\newblock {\em Transactions of the Association for Computational Linguistics\/}
  4:313--327.

\bibitem[{Kuncoro et~al.(2016)Kuncoro, Ballesteros, Kong, Dyer, and
  Smith}]{kuncoro2016distilling}
Adhiguna Kuncoro, Miguel Ballesteros, Lingpeng Kong, Chris Dyer, and Noah~A.
  Smith. 2016.
\newblock Distilling an ensemble of greedy dependency parsers into one mst
  parser.
\newblock In {\em EMNLP\/}.

\bibitem[{Lei et~al.(2015)Lei, Barzilay, and Jaakkola}]{lei2015molding}
Tao Lei, Regina Barzilay, and Tommi Jaakkola. 2015.
\newblock Molding cnns for text: non-linear, non-consecutive convolutions.
\newblock {\em Empirical Methods in Natural Language Processing\/} .

\bibitem[{Ma and Hovy(2017)}]{ma2017neural}
Xuezhe Ma and Eduard Hovy. 2017.
\newblock Neural probabilistic model for non-projective mst parsing.
\newblock {\em arXiv preprint: 1701.00874\/} .

\bibitem[{McDonald et~al.(2005)McDonald, Pereira, Ribarov, and
  Hajic}]{mcdonal2005non}
Ryan McDonald, Fernando Pereira, Kiril Ribarov, and Jan Hajic. 2005.
\newblock Non-projective dependency parsing using spanning tree algorithms.
\newblock In {\em Proc. Human Language Technology Conf. and Conf. Empirical
  Methods Natural Language Process. (HLT/EMNLP)\/}. pages 523--530.

\bibitem[{Nivre(2009)}]{nivre2009non}
Joakim Nivre. 2009.
\newblock Non-projective dependency parsing in expected linear time.
\newblock In {\em Proceedings of the 47th Annual Meeting of the ACL and the 4th
  IJCNLP of the AFNLP\/}.

\bibitem[{Srivastava et~al.(2014)Srivastava, Hinton, Krizhevsky, Sutskever, and
  Salakhutdinov}]{srivastava2014dropout}
Nitish Srivastava, Geoffrey~E Hinton, Alex Krizhevsky, Ilya Sutskever, and
  Ruslan Salakhutdinov. 2014.
\newblock Dropout: a simple way to prevent neural networks from overfitting.
\newblock {\em Journal of Machine Learning Research\/} 15(1):1929--1958.

\bibitem[{Strubell et~al.(2017)Strubell, Verga, Belanger, and
  McCallum}]{strubell2017fast}
Emma Strubell, Patrick Verga, David Belanger, and Andrew McCallum. 2017.
\newblock Fast and accurate sequence labeling with iterated dilated
  convolutions.
\newblock {\em arXiv preprint: arXiv:1702.02098\/} .

\bibitem[{Toutanova et~al.(2015)Toutanova, Chen, Pantel, Poon, Choudhury, and
  Gamon}]{toutanova2015representing}
Kristina Toutanova, Danqi Chen, Patrick Pantel, Hoifung Poon, Pallavi
  Choudhury, and Michael Gamon. 2015.
\newblock Representing text for joint embedding of text and knowledge bases.
\newblock In {\em Proceedings of the 2015 Conference on Empirical Methods in
  Natural Language Processing\/}. Association for Computational Linguistics,
  pages 1499--1509.

\bibitem[{Tutte(1984)}]{tutte1984graph}
William~Thomas Tutte. 1984.
\newblock {\em Graph theory\/}, volume~11.
\newblock Addison-Wesley Menlo Park.

\bibitem[{van~den Oord et~al.(2016)van~den Oord, Dieleman, Zen, Simonyan,
  Vinyals, Graves, Kalchbrenner, Senior, and
  Kavukcuoglu}]{vandenoord2016wavenet}
Aaron van~den Oord, Sander Dieleman, Heiga Zen, Karen Simonyan, Oriol Vinyals,
  Alex Graves, Nal Kalchbrenner, Andrew Senior, and Koray Kavukcuoglu. 2016.
\newblock Wavenet: A generative model for raw audio.
\newblock {\em arXiv preprint arXiv:1609.03499\/} .

\bibitem[{Weiss et~al.(2015)Weiss, Alberti, Collins, and
  Petrov}]{weiss2015structured}
David Weiss, Chris Alberti, Michael Collins, and Slav Petrov. 2015.
\newblock Structured training for neural network transition-based parsing.
\newblock In {\em Annual Meeting of the Association for Computational
  Linguistics\/}.

\bibitem[{Yu and Koltun(2016)}]{yu2015multi}
Fisher Yu and Vladlen Koltun. 2016.
\newblock Multi-scale context aggregation by dilated convolutions.
\newblock In {\em International Conference on Learning Representations
  (ICLR)\/}.

\bibitem[{Zhang et~al.(2015)Zhang, Zhao, and LeCun}]{zhang2015character}
Xiang Zhang, Junbo Zhao, and Yann LeCun. 2015.
\newblock Character-level convolutional networks for text classification.
\newblock In {\em Advances in Neural Information Processing Systems 28
  (NIPS)\/}.

\end{thebibliography}
\bibliographystyle{emnlp_natbib}

\end{document}